\documentclass[sigconf]{acmart}

\usepackage{adjustbox}
\AtBeginDocument{%
  \providecommand\BibTeX{{%
    \normalfont B\kern-0.5em{\scshape i\kern-0.25em b}\kern-0.8em\TeX}}}

\copyrightyear{2023} 
\acmYear{2023} 
\setcopyright{rightsretained} 
\acmConference[ICPE '23]{Proceedings of the 2023 ACM/SPEC International Conference on Performance Engineering}{April 15--19, 2023}{Coimbra, Portugal}
\acmBooktitle{Proceedings of the 2023 ACM/SPEC International Conference on Performance Engineering (ICPE '23), April 15--19, 2023, Coimbra, Portugal}
\acmDOI{10.1145/3578244.3583731}
\acmISBN{979-8-4007-0068-2/23/04}

\begin{document}

\title{Predicting the Performance of a Computing System with Deep Networks}

\author{Mehmet Cengiz}
\affiliation{%
\institution{Newcastle University}
\streetaddress{Address}
\city{Newcastle upon Tyne}
\country{UK}}
\email{m.cengiz2@ncl.ac.uk}

\author{Matthew Forshaw}
\affiliation{%
\institution{Newcastle University}
\streetaddress{Address}
\city{Newcastle upon Tyne}
\country{UK}}
\email{matthew.forshaw@ncl.ac.uk}

\author{Amir Atapour-Abarghouei}
\affiliation{%
\institution{Durham University}
\streetaddress{Address}
\city{Durham}
\country{UK}}
\email{amir.atapour-abarghouei@durham.ac.uk}

\author{Andrew Stephen McGough}
\affiliation{%
\institution{Newcastle University}
\streetaddress{Address}
\city{Newcastle upon Tyne}
\country{UK}}
\email{stephen.mcgough@newcastle.ac.uk}

\renewcommand{\shortauthors}{Cengiz, Forshaw, Atapour-Abarghouei, McGough}

\begin{abstract}
Predicting the performance and energy consumption of computing hardware is critical for many modern applications. This will inform procurement decisions, deployment decisions, and autonomic scaling. Existing approaches to understanding the performance of hardware largely focus around benchmarking -- leveraging standardised workloads which seek to be representative of an end-user’s needs. Two key challenges are present; benchmark workloads may not be representative of an end-user’s workload, and benchmark scores are not easily obtained for all hardware. Within this paper, we demonstrate the potential to build Deep Learning models to predict benchmark scores for unseen hardware. We undertake our evaluation with the openly available SPEC 2017 benchmark results. We evaluate three different networks, one fully-connected network along with two Convolutional Neural Networks (one bespoke and one ResNet inspired) and demonstrate impressive $R^2$ scores of 0.96, 0.98 and 0.94 respectively.
\end{abstract}



\received{20 February 2007}
\received[revised]{12 March 2009}
\received[accepted]{5 June 2009}

\maketitle

\section{Introduction}
Performance benchmarks are commonly used as a tool to better understand systems. This includes informing procurement decisions and, through the operation of systems, to inform deployment and scaling decisions. These benchmarks seek to understand the likely performance of a user's workload, but also energy consumption. While benchmarks show good potential to gain an understanding of performance, it is typically infeasible to benchmark all possible combinations of workload and hardware. This problem is exacerbated in environments which exhibit hardware heterogeneity.

Benchmarks~\cite{anand2008benchmarking, camp1989benchmarking, maire2005typology} -- which produce metrics~\cite{cam_bench, umd} on different hardware under specific workloads, help to identify the `best' hardware. The metrics can then be compared for different hardware options, supporting judgements as to how a specific user's workload would be expected to perform. 

We seek to resolve the challenge of evaluating the performance for previously unseen hardware-workload combinations, using the SPEC CPU 2017 dataset. 
Previous efforts using linear regression (e.g.,~\cite{benjamin_spec}) have demonstrated the potential to predict performance metrics, but perform poorly for non-linear aspects of hardware evolution. In our work we present a data cleaning pipeline to ensure the data is amenable to modelling.

We explore the potential of three Deep Networks to better model non-linear relationships in the benchmark data. We evaluate a number of fully-connected networks (often referred to as multilayer perceptrons (MLP)) due to the tabular format of the dataset as well as Convolutional Neural Networks (CNN). Originally developed for learning from image-based data (2-dimensional, greyscale, or 3-dimensional, colour), CNNs have recently gained traction in the case of 1-dimensional datasets such as tables~\cite{tab_conv, Butur, kaggle_1d}. For the first CNN approach, we evaluate a number of networks which contain convolution and pooling operations whilst for the second CNN approach, we evaluate adding residual blocks as proposed in ResNet~\cite{resnet}. We perform a hyperparameter tuning process within each of these networks. This allows us to demonstrate our approach can accurately predict unseen benchmark results. From this we are able to achieve $R^2$ scores of 0.96, 0.98 and 0.94 respectively, compared to 0.53 for linear regression.

\begin{table*}[t!h!]
  \caption{An overview of the prediction studies that used SPEC datasets.}
  \label{tab:rw}
  \begin{tabular}{cp{4cm}p{5cm}p{6cm}}
    \toprule
    Work&Dataset(s)&Technique(s)&Prediction \\
    \midrule
    \cite{benjamin_spec}&
        SPEC CPU / SPEC Java Server&
        Custom linear regression model&
        Server benchmark performances \\
    \cite{berkin2008}&
        SPEC 2006&
        Custom linear regression model&
        Performance of future systems \\ 
    \cite{eyerman_mech}&
        SPEC CPU2000 / CPU2006&
        Hybrid mechanistic-empirical model&
        Commercial processor performance \\
    \cite{Jiang}&
        SPEC OpenMP&
        Classic fractal-based sampling&
        Accelerating multithreaded app simulation \\
    \cite{Zheng_spec}&
        SPEC 2006&
        Fine-grained phase-based approach&
        Performance and power \\
    \cite{lopez2018}&
        SPEC 2017&
        Multiple Neural Networks&
        Computer hardware configuration \\
    \cite{tousi}&
        SPEC 2017&
        Multi-layer perceptron&
        Computer performance \\
    \textbf{Ours}&
        SPEC 2017&
        MLP, CNN&
        Computer performance \\
  \bottomrule
\end{tabular}
\end{table*}

The remainder of this paper is organised as follows. In Section \ref{related}, we discuss prior work focusing on performance prediction. We outline our methodology in Section \ref{method}. We present our results in Section \ref{results} and explore Threats to Validity in Section~\ref{ttv}. We conclude and outline areas of future work in Section \ref{conc}.


\section{Related Work}
\label{related}

Here we present prior work on ML-based performance prediction of computer systems.

Performance prediction is the process of predicting some performance metric for a system based on known characteristics of that system, which is sometimes referred to as empirical performance modeling~\cite{emp_per}. However, we will reduce the scope of our study here down to the prediction of performance metrics for computer systems. In general, performance prediction is for values which can take any value within a given range (e.g., time to complete some task or a numeric value used to compare different systems). As such, the work here focuses on regression techniques.

 One of the earliest studies was performed by Ein-Dor and Feld-messer~\cite{ein-dor}. They claimed that by using readily available data on CPU characteristics, it is possible to predict a given CPU performance. However, their work is based around simple statistical approaches and cannot be used for the SPEC performance predictions we wish to perform here. {I}pek \textit{et al.}~\cite{engin} used artificial neural networks to predict Instructions per Cycle (IPC) of a given system. Their dataset contains L1 and L2 cache sizes --the first and second caches in the hierarchy of cache levels-- and front-side bus bandwidth. Their experiments showed that their model predicts IPC with only a 1-2\% error.

Li \textit{et al.}~\cite{li_cloud} carried the empirical performance prediction domain to the cloud environment by developing a tool named CloudProphet. This was effectively a trace-and-replay tool to predict a legacy application's performance if migrated to a cloud infrastructure. As our work here focuses on prediction of benchmark scores, this would not be easily translatable to our work, though it could form a good starting point for predicting the performance of a specific workload on another (non-cloud) computer. 

Upadhyay \textit{et al.}~\cite{Upadhyay} discuss performance prediction issues from a different point of view. Their motivation is to consider the other components of a systems hardware while designing a CPU. For selecting the best combination of CPU, they used data mining techniques. Although this could be applied to the SPEC datasets we would argue that the non-linear nature of new hardware would make this a less than accurate approach.

A number of prediction approaches have been proposed for prediction of performance metrics for GPUs. Ardalani \textit{et al.}~\cite{ardalani} focused on GPU performances and designed an ensemble of regression learners named Cross-Architecture Performance Prediction (XAPP). However, they intended to predict GPU performances using single-threaded CPU implementations. They achieved a 26.9\% average error on a set of 24 real-world kernels. As they mentioned in their paper, their study cannot capture the impact of texture memory and constant memory. On the other hand, adhering to their implication, this is the problem of having a small dataset that contains 122 data points. Therefore, since our dataset contains more than 20K data points, we require more sophisticated models.

The work by Justus \textit{et al.}~\cite{justus} forms inspiration for our work as they used Multi-Layer Perceptrons for the prediction of execution time for training Deep Learning networks. However, we take this work further by using Convolutional Neural Networks for our predictions and apply it to the SPEC dataset.

\subsection{Predictions from the SPEC datasets}
A number of works have addressed the problem of predicting metrics for the SPEC datasets. As there have only been two prior works which address the SPEC 2017 dataset, we expand our discussion here to cover all of the SPEC datasets. A summary of these works can be found in Table~\ref{tab:rw}.

Lee~\cite{benjamin_spec} and Ozisikyilmaz \textit{et al.}~\cite{berkin2008} used linear regression models for predicting benchmark performance. Our work seeks to overcome potential limitations by modelling non-linear responses.

Eyerman \textit{et al.}~\cite{eyerman_mech} developed a mechanistic model built on interval analysis which breaks the total execution time into intervals based on missed events, for out-of-order superscalar processors.

Jiang \textit{et al.}.~\cite{Jiang} presented a study to evaluate design alternatives for computer architectures. They designed a fractal-based sampling to speed up parallel microarchitecture simulation with multithreaded applications. Due to the fact that they mainly intend to obtain samples from parallel programming datasets, the only similarity with our study is the use of SPEC-based datasets.

Zheng \textit{et al.}~\cite{Zheng_spec} proposed a unified learning-based framework named LACross to estimate time-varying software performance and power consumption on a target hardware platform. 

Lopez \textit{et al.}~\cite{lopez2018} used multiple neural networks for a classification task for predicting the best computer hardware configuration options. Although their work demonstrates the validity of using Deep Learning on SPEC datasets, their underlying problem is quite different to ours. The closest work to ours is that of Tousi and Lujan~\cite{tousi}, which uses MLPs for the prediction of computer performance. We go further by demonstrating how the use of Convolutional Neural Networks can be used to provide better results.

\section{Methodology}
\label{method}
All experiments are run on a Tesla T4 GPU and two Intel Xeon(R) CPUs @ 2.30GHz, and 12 GB of memory. As the SPEC 2017 dataset is not directly in a format which can be used for machine learning, we first discuss the process used for dataset cleansing in order to provide data which can be fed directly to our Deep Learning networks. We then go on to cover the search space of Deep Learning networks which we have evaluated as part of this work.

\begin{table}[t]
  \caption{Columns of SPEC2017}
  \label{tab:spec_columns}
  \begin{tabular}{cl}
    \toprule
    Data Type&Column \\
    \midrule
       String  & Benchmark, Hardware Vendor, System,  \\
        & Processor, CPU(s) Orderable, 1st Level Cache, \\ 
        &
             2nd Level Cache, 3rd Level Cache, Other Cache, \\
        &    Storage, Operating System, File System,  \\
        & Compiler, License, Tested By, Test Sponsor \\ \hline
    Numerical & Peak Result, Base Result, Energy Peak Result, \\
              & Energy Base Result, \# Cores, \# Chips,  Memory, \\
              & \# Enabled Threads Per Core, Processor MHz  \\ \hline
    Binary & Parallel \\ \hline
    Ternary & Base Pointer Size \\ \hline
    Quaternary & Peak Pointer Size \\ \hline
    Date & HW Avail, SW Avail, Test Date, Published, \\
    (mon-yyyy) &  Updated \\ \hline
    Text & Disclosures \\
  \bottomrule
\end{tabular}
\end{table}


\subsection{Dataset cleansing}
\label{subse:data-clensing}
Within this work we consider how to prepare SPEC  2017 benchmark dataset for machine learning. The dataset includes 34 attributes as illustrated in Table~\ref{tab:spec_columns}. The numeric columns \emph{Peak Result} and \emph{Base Result} represent the response time of systems under load or no load respectively and are the values we seek to predict in this work. We perform the following pre-processing on the data, making it amenable for model training. Our approach to mitigate inconsistencies and data quality issues include the following:

\begin{description}
	\item[Alphanumeric cleaning:] Non-alphanumeric characters such as tabs and escape characters are removed from the dataset. We also remove spaces from column names to make downstream processing easier. All characters are converted to lower case to remove inconsistencies.
	\item[Removal of outliers:] Some of the \emph{Base Result} values were zero, which is clearly incorrect. As there were only a small number of these, they are removed.
        \item[Making units consistent:] Units varied across the data (e.g., memory in KB, MB, GB). All units are standardised to MB.
	\item[Make columns categorical:] Many of the columns although appearing to allow arbitrary data are actually highly constrained (e.g., Memory can only take a small range of values). As such, the set of these values was determined and the data was replaced with categorical labels.
	\item[Removal of highly correlated columns:] We used Kendall's rank correlation~\cite{kendall} to identify those columns which are highly correlated. It was determined, in our case, that the columns `CPU(s) Orderable',  `Energy Base Result',  `License',  `Parallel',  `System',  `Test Sponsor', and `Tested By' were more than 70\% correlated with other columns. As strongly correlated variables may have almost the same ability to predict the result value for observation, due to their linear dependence, they were eliminated. It should be noted that we also evaluated Pearson and Spearman correlation and obtained similar results.
\end{description}

\subsection{Searching for the `best' Neural Network}
The shape (layers and neurons per layer) of Deep Learning networks significantly impact performance. We perform a space search for the `best' network for the SPEC data. We identify three network structures, two trapezium and one rectangular and populate these with either single neurons, Convolutional nodes or Residual Nodes.\label{archSearch} We evaluate three core network designs within this work. Those of fully-connected networks, convolutional neural networks and networks which use Residual blocks as proposed by the ResNet architecture~\cite{resnet}. We detail the design of each of these architectures:

\subsubsection{Fully-Connected Networks:}
\begin{figure}[b]
  \centering
  \includegraphics[width=\linewidth]{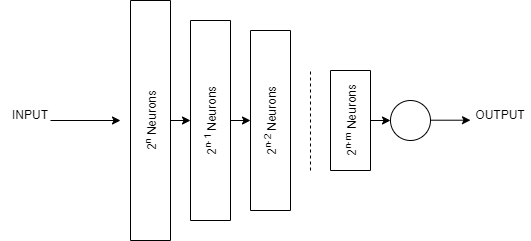}
  \caption{Sketch view of trapezium-shaped MLPs}
  \label{fig:rMLP}
  \Description{Sketch view of trapezium-shaped MLPs}
\end{figure}

We evaluate three network structures, those of a strictly decreasing number of neurons per layer shaped network -- which we will refer to as a trapezium network hereafter, see Figure \ref{fig:rMLP}, the reverse of this -- referred to as a reverse trapezium -- and that of a rectangular network with the same number of neurons in each layer. For the trapezium network the first layer has $2^n$ neurons. Each subsequent layer has half the number of neurons as the previous layer. The penultimate layer has $2^{n-m}$ neurons where $n-m > 1$. We vary the values of $n$ in the range $[4, ..., 11]$ and $m$ in the range $[1, ..., 10]$. The final layer of the network contains just a single neuron to provide the regression result. Reverse trapezium networks flip the order of the layers (apart from the last) having the narrowest layer first and the widest layer last.

The rectangular networks contain $m$ layers and have $2^n$ neurons in each layer, with a final layer containing only one neuron to provide the regression result. Although this network does not vary in shape between layers, the network learns weights which cause some neurons in a level to become redundant, effectively learning itself the number of neurons to place in each layer.

\subsubsection{CNN design:}
\begin{figure}
    \centering
    \includegraphics[width=\linewidth]{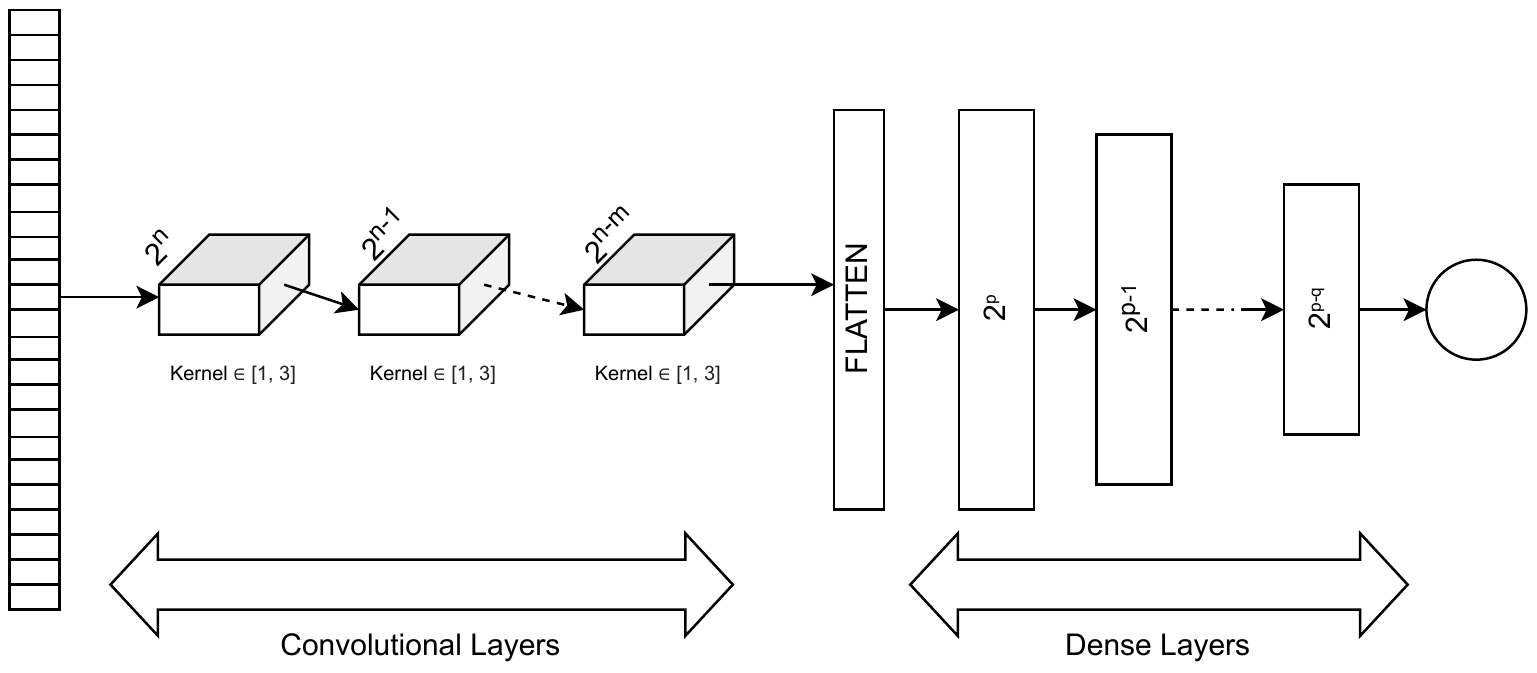}
    \caption{The CNN network structure}
    \label{fig:tri_cnn}
\end{figure}

The CNN network consists of a number of convolutional layers followed by a fully-connected set of layers. Figure \ref{fig:tri_cnn} illustrates the shape of these networks. It should be noted that in these cases the fully-connected layers are smaller than those where we only use fully-connected layers. As our data is tabular, we use 1D convolutional layers -- i.e. our kernels/filters are 1D and of size $k \in [2,...,5]$. Again, we adopt the trapezium format of the first convolutional layer having a width (number of filters) of $2^n$ and each subsequent layer having half the width of the previous layer. With the last convolutional layer having a width of $2^{n-m}$ ($n-m > 1$). The fully-connected layers are trapezium in shape and range in nodes per layer between $2^p$ and $2^{p-q}$. We allow $n$, $m$, $p$ and $q$ to vary in the ranges $[7, ..., 11], [4, ..., 7], [7, ..., 11],$ and $[5, ..., 7]$, respectively. Initial experiments indicated that searches within these ranges yielded the best results.

\subsubsection{Residual design:}
\begin{figure}[b]
  \centering
  \includegraphics[width=\linewidth]{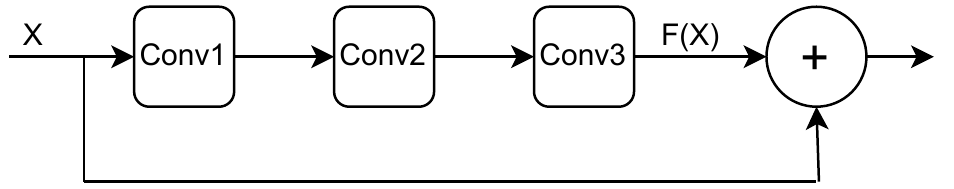}
  \caption{The identity block}
  \label{fig:IdentBlock}
  \Description{A block with three convolutional layers with different specifications.}
\end{figure}

We adopt Residual blocks ~\cite{resnet}, where a `bypass' link around a set of convolutional units is merged with the output from the convolutional units. Figure \ref{fig:IdentBlock} illustrates this network topology and we refer to this hereafter as the identity block. The width of the input and output to the identity block must be the same ($2^p$). By convention, the kernel size of the first two convolutions are $2^{p-2}$ with the kernel size of the last convolution being $2^p$ to restore the original size. We allow $p \in [6, ..., 11]$.

\begin{figure}[t]
  \centering
  \includegraphics[width=\linewidth]{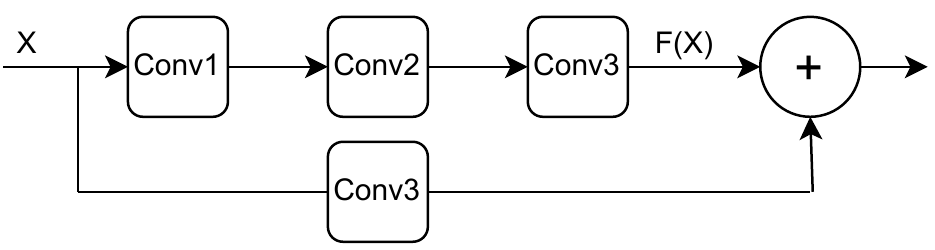}
  \caption{The convolutional block}
  \label{fig:ConvBlock}
  \Description{A block with three convolutional layers with different specifications and a skipping block with the same size as the third layer.}
\end{figure}

One restriction of the original identity block is that the shape of the data entering the block must be the same as the shape of the output -- otherwise the merging of the data from the `bypass' will not be possible. In order to overcome this, we use a convolution unit to the `bypass' which has the same output width as the final convolution in the main path -- see Figure \ref{fig:ConvBlock}. In this case, the first two convolutions on the main path have a width of $2^{p-2}$, while the last convolution on the main path and the `bypass' path have widths of $2^{p}$. We refer to this as a convolutional block. 

\begin{figure}
    \centering
    \includegraphics[width=\linewidth]{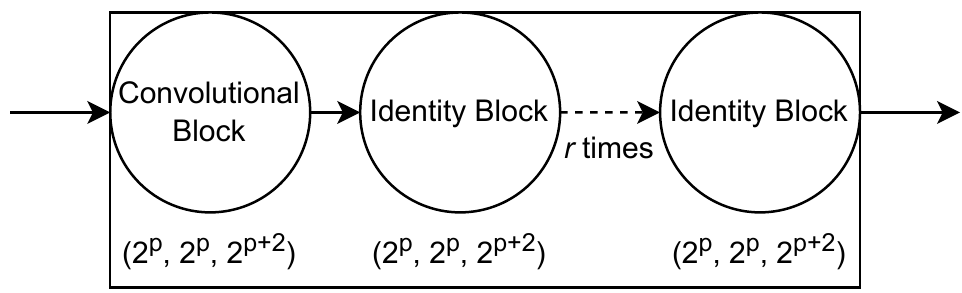}
    \caption{A superblock constructed from a convolution block and $r$ identity blocks}
    \label{fig:superblock}
\end{figure}

The two block templates are then combined to produce a superblock (Figure \ref{fig:superblock}). Each superblock starts with a convolutional block followed by $r$ identity blocks. The width of the output for each block (both identity and convolution) within a block will be $2^{p}$, also the output width of the whole superblock.

\begin{figure}[h]
  \centering
  \includegraphics[width=\linewidth]{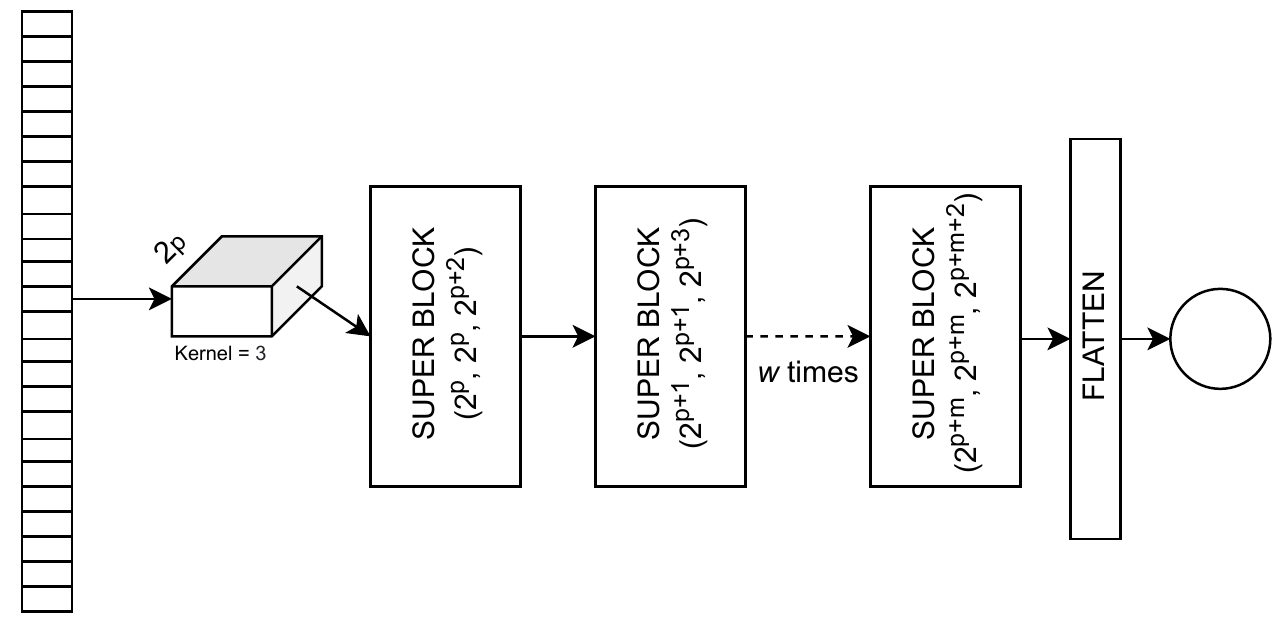}
  \caption{The ultimate design of our ResNet model}
  \label{fig:ResNet}
  \Description{1D ResNet.}
\end{figure}

Superblocks can then be concatenated together as in Figure \ref{fig:ResNet}. Here, the original vector data is fed into a set of $w$ superblocks. Following the convention of ResNet, the width of output from each superblock will be double that of the previous superblock. Finally the output from the last superblock will be flattened before being fed into a single neuron to predict the regression value.

\subsection{Hyperparameter search}
In addition to performing an neural architecture search over the architecture range specified in \ref{archSearch}, we also conducted a thorough search across the hyperparameters which could be used for the networks. This included the optimiser, the number of training epochs, the loss function and the activation function. 

\subsubsection{Optimiser:} The optimiser is used to determine how the wei-ghts of the network are updated after each training step. This work focuses on three of the most commonly used optimisers:

{\bf SGD:} Stochastic Gradient Decent is the original optimiser used for Deep Learning. Although strictly speaking Gradient decent performs an update after each training sample is processed, we adopt the normal convention of performing the optimisation step after each batch of data is processed -- more correctly referred to as Batched Stochastic Gradient Decent.

{\bf RMSprop:} Root Mean Squared Propagation extends SGD by applying decaying average partial gradients to the step size of each parameter. The optimiser focuses more on recent gradients.

{\bf Adam:} ADAptive Moment estimation~\cite{adam} is an extension of SGD. Like RMSprop, Adam adopts a separate learning rate for each parameter. While RMSprop uses the average of the first moment, Adam also uses the average of the second moment when choosing how to adapt the learning rates.

\subsubsection{Loss function:} The loss function is used to determine the difference between the predicted values and the true values. We evaluate two loss functions, Mean Squared Error (MSE, Equation~\ref{eqn:mse}) and Mean Absolute Error (MAE, Equation~\ref{eqn:mae}),

\hspace{-0.6cm}\begin{minipage}{.5\linewidth}
\begin{equation}\label{eqn:mse}
    MSE = \frac{1}{N}\sum_{i=1}^N (y_i' - y_i)^2 ,
\end{equation}
\end{minipage}
\begin{minipage}{.5\linewidth}
\begin{equation}\label{eqn:mae}
    MAE = \frac{1}{N}\sum_{i=1}^N |y_i' - y| ,
\end{equation}
\end{minipage}

\noindent where number of samples is $N$, $y_i'$ and $y_i$ are predicted and true values respectively. Larger errors will have a larger impact on MSE's loss value which we would expect to lead to fewer outlier values.

\subsubsection{Activation function:}

The activation function provides the non-linear element within the networks. We evaluated three commonly used activation functions within our work: sigmoid, tanh and ReLU. For many problems ReLU has been shown to be the most effective activation function. However, there are a number of cases where the other activation functions are more suited to the problem at hand. As is the convention, no activation function was used on the final output layer to allow for arbitrary output values.

\subsubsection{Stride:}
Stride is the number of cells that shifts over the input matrix. While stride size is adaptive in the CNN experiments, we kept it fixed in the residual-inspired models except for the starting layer of the convolutional blocks. Our dataset has 24 columns for independent variables, making the shape of input data (1, 24), we defined the stride size in the range of $[1, ..., 4]$.

\subsection{Implementation Details}

We use an 80-20 training-test split. Further, the training data is split into training and validation sets as 80\% and 20\% respectively. The batch size is determined as 10. Each model is trained 5 times with different random seeds to obtain its average performance. The data splitting process was purposely designed to demonstrate the real-world future unseen data in operation, and adjusting class distributions via any controlled split approach such as cross-validation would go against an uncontrolled future configuration of data~\cite{cv}. By using random seeds to generate random splits of data for each experiment, we can be confident that our models are capable of responding to random distributions of data. in the real world.

We use the Glorot uniform initialiser~\cite{glorot} for initialising the parameters within our networks, which sets the weights so they are equal across all layers in terms of the variance of the activations. The gradient is kept from exploding or vanishing by the constant variance. In addition, the initial bias value(s) were set to zero~\cite{kerasInit}.

We allowed the number of epochs to vary between 100 and 300 in steps of 50. We stopped training after 300 epochs, where models demonstrated optimal performance on the test set.

\subsubsection{Baseline Models}
We consider three baseline models:

{\bf Linear Regression:} We would expect that this model would perform well for similar hardware, but perform poorly when there is a non-linear change in hardware performance.

{\bf Support Vector Regression:} SVR is often better than a linear regression model as it is able to fit better to the model. However, it still suffers from the fact that it is a linear model and hence is not expected to adapt well to step-changes in the hardware.

{\bf Random Forest Regression:} This is an ensemble technique which does not suffer from the linear model problems of the other two approaches. It does, however, require prior examples of hardware types to be able to predict new hardware accurately. We would therefore expect this to be better than the other baseline models, but less likely to be adaptable as the Deep Learning models.

\subsubsection{Evaluation Metrics:} We evaluate model performance using MSE (Equation \ref{eqn:mse}), MAE (Equation \ref{eqn:mae}) and $R^2$:
$$
  R^2 = 1 - \frac{\sum_{i=1}^n (y_i - \hat{y_i})^2}{\sum_{i=1}^n (y_i - \bar{y})^2},
$$
where $y_i$ is the true value, $\hat{y_i}$ is the predicted value and $\bar{y}$ is the mean of all true values. For both the $R^2$ and MSE, values further from their predicted value are going to have a more significant impact on the results. In order to measure the models' vulnerability to outliers, focusing on MSE would be preferable. A focus on $R^2$ would reduce outliers at the expense of overall accuracy.
\section{Results}
\label{results}

\begin{table*}[ht]
    \caption{The results of the best deep networks and machine learning models -- Order of $R^2$}
    \label{tab:best20}
    \resizebox{2.1\columnwidth}{!}{
        \begin{tabular}{cccccccccccc}
            \toprule
                \# & Architecture & Loss Fn & Kernel Sizes & Stride Sizes & Number of Filters (m, n) & Neurons in Layers (p, q) & Optimizer & Epochs & R2 & MAE & MSE \\
            \midrule
                1      & TriCNN  & MAE & 3 & 1 & (9, 7)          & {[}9, …, 5{]}  & Adam      & 250 & 0.98638701  & 5.67389728  & 465.3285655 \\
                2      & TriCNN  & MAE & 3 & 2 & (9, 7, 6, 5, 4) & {[}9, …, 4{]}  & Adam      & 250 & 0.98590661  & 5.83946465  & 476.0394343 \\
                3      & TriCNN  & MAE & 3 & 2 & (9, 7)          & {[}9, …, 5{]}  & Adam      & 300 & 0.98579341  & 5.76197731  & 494.124225 \\
                4      & TriCNN  & MAE & 3 & 1 & (9, 7)          & {[}9, …, 5{]}  & Adam      & 150 & 0.98529142  & 6.25318407  & 513.9629513 \\
                5      & TriCNN  & MAE & 3 & 2 & (9, 7, 6, 5, 4) & {[}9, …, 4{]}  & RmsProp   & 150 & 0.98282719  & 7.14056732  & 620.2982421 \\
                6      & TriCNN  & MAE & 3 & 2 & (9, 7, 6, 5, 4) & {[}9, …, 4{]}  & Adam      & 200 & 0.98280914  & 6.03564805  & 582.3068145 \\
                7      & TriCNN  & MAE & 3 & 2 & (9, 7, 6, 5, 4) & {[}9, …, 4{]}  & Adam      & 300 & 0.98278342  & 5.61076184  & 582.0247239 \\
                8      & TriCNN  & MAE & 3 & 1 & (9, 7)          & {[}9, …, 5{]}  & Adam      & 300 & 0.98107176  & 5.78137347  & \multicolumn{1}{l}{645.4129883} \\
                9      & TriCNN  & MAE & 3 & 2 & (9, 7)          & {[}9, …, 5{]}  & RmsProp   & 250 & 0.98095925  & 6.72097815  & 669.8856237 \\
                10     & TriCNN  & MAE & 3 & 1 & (9, 7)          & {[}9, …, 5{]}  & Adam      & 200 & 0.98089907  & 6.32291809  & 665.1641919 \\
                11     & TriCNN  & MAE & 3 & 2 & (9, 7)          & {[}9, …, 5{]}  & Adam      & 150 & 0.98047251  & 6.71537772  & 663.7030719 \\
                12     & TriCNN  & MAE & 3 & 1 & (7, 6, 5, 4)    & {[}9, …, 5{]}  & RmsProp   & 300 & 0.98038864  & 6.9974749   & \multicolumn{1}{l}{653.5821786} \\
                $\sim$ & RF      &     &   &   &                 &                &           &     & 0.9803076   & 4.76701531  & 688.0001262 \\
                13     & TriCNN  & MAE & 3 & 1 & (7, 6, 5, 4)    & {[}9, …, 5{]}  & RmsProp   & 200 & 0.98002879  & 7.62788323  & 684.7595471 \\
                14     & TriCNN  & MAE & 2 & 1 & (9, 7)          & {[}11, …, 6{]} & Adam      & 150 & 0.9793459   & 6.519971    & 703.0615545 \\
                15     & TriCNN  & MAE & 3 & 2 & (9, 7)          & {[}9, …, 5{]}  & Adam      & 100 & \multicolumn{1}{l}{0.97782539} & \multicolumn{1}{l}{8.23651529} & 754.5381605 \\
                16     & TriCNN  & MAE & 3 & 2 & (9, 7, 6, 5, 4) & {[}9, …, 4{]}  & Adam      & 100 & 0.97748578  & 7.30871799  & 757.4994833 \\
                17     & TriCNN  & MAE & 3 & 2 & (9, 7, 6, 5, 4) & {[}9, …, 4{]}  & Adam      & 150 & 0.97726148  & 6.65855022  & 772.0747562 \\
                18     & TriCNN  & MAE & 3 & 1 & (7, 6, 5, 4)    & {[}9, …, 5{]}  & RmsProp   & 250 & 0.97665471  & 7.86703389  & \multicolumn{1}{l}{775.8960386} \\
                19     & TriCNN  & MAE & 3 & 2 & (9, 7, 6, 5, 4) & {[}9, …, 4{]}  & RmsProp   & 250 & 0.97650919  & 7.97325412  & 852.3545636 \\
                20     & TriCNN  & MAE & 3 & 2 & (9, 7)          & {[}9, …, 5{]}  & RmsProp   & 300 & 0.97636563  & 6.91501173  & 816.7881606 \\
                45     & TriMLP  & MAE &   &   &                 & {[}11, …, 6{]} & Adam      & 250 & \multicolumn{1}{l}{0.97347275} & \multicolumn{1}{l}{9.12443258} & \multicolumn{1}{l}{906.1439402} \\
                159    & Residual& MAE & \multicolumn{2}{l}{Number of Superblocks =   (2, 5, 5, 2)} & \multicolumn{1}{l}{((6, 6, 8), (7, 7, 9), (8, 8, 10), (9, 9, 11))} & 1 & RmsProp   & 250 & 0.95007233 & 10.595069 & 1006.134564 \\
                $\sim$ & LR      &     &   &   &                 &                &           &     & 0.52639158  & 82.4596122  & 15761.16107 \\
                $\sim$ & SVR     &     &   &   &                 &                &           &     & -0.0045634  & 113.749207  & 33448.30886 \\
            \bottomrule
        \end{tabular}
    }
    \resizebox{1.5\columnwidth}{!}{* TriCNN = Trapezium-shaped CNN, RF = Random Forest Regression, TriMLP = Trapezium-shaped MPL, LR = Linear Regression, SVR = Support Vector Regression}
\end{table*}

\begin{table*}[ht]
    \caption{The results of the best deep networks and machine learning models -- Order of MSE}
    \label{tab:best20mse}
    \resizebox{2.1\columnwidth}{!}{
        \begin{tabular}{cccccccccccc}
            \toprule
                 \# & Architecture & Loss Fn & Kernel Sizes & Stride Sizes & Number of Filters (m, n) & Neurons in Layers (p, q) & Optimizer & Epochs & R2 & MAE & MSE \\
            \midrule
                1      & TriCNN   & MAE & 3 & 1 & (9, 7)          & {[}9, …, 5{]}  & Adam      & 250    & 0.98638701 & 5.67389728 & 465.3285655 \\
                2      & TriCNN   & MAE & 3 & 2 & (9, 7, 6, 5, 4) & {[}9, …, 4{]}  & Adam      & 250    & 0.98590661 & 5.83946465 & 476.0394343 \\
                3      & TriCNN   & MAE & 3 & 2 & (9, 7)          & {[}9, …, 5{]}  & Adam      & 300    & 0.98579341 & 5.76197731 & 494.124225 \\
                4      & TriCNN   & MAE & 3 & 1 & (9, 7)          & {[}9, …, 5{]}  & Adam      & 150    & 0.98529142 & 6.25318407 & 513.9629513 \\
                5      & TriCNN   & MAE & 3 & 2 & (9, 7, 6, 5, 4) & {[}9, …, 4{]}  & Adam      & 300    & 0.98278342 & 5.61076184 & 582.0247239 \\
                6      & TriCNN   & MAE & 3 & 2 & (9, 7, 6, 5, 4) & {[}9, …, 4{]}  & Adam      & 200    & 0.98280914 & 6.03564805 & 582.3068145 \\
                7      & TriCNN   & MAE & 3 & 2 & (9, 7, 6, 5, 4) & {[}9, …, 4{]}  & RmsProp   & 150    & 0.98282719 & 7.14056732 & 620.2982421 \\
                8      & TriCNN   & MAE & 3 & 1 & (9, 7)          & {[}9, …, 5{]}  & Adam      & 300    & 0.98107176 & 5.78137347 & \multicolumn{1}{l}{645.4129883} \\
                9      & TriCNN   & MAE & 3 & 1 & (7, 6, 5, 4)    & {[}9, …, 5{]}  & RmsProp   & 300    & 0.98038864 & 6.9974749  & \multicolumn{1}{l}{653.5821786} \\
                10     & TriCNN   & MAE & 3 & 2 & (9, 7)          & {[}9, …, 5{]}  & Adam      & 150    & 0.98047251 & 6.71537772 & 663.7030719 \\
                11     & TriCNN   & MAE & 3 & 1 & (9, 7)          & {[}9, …, 5{]}  & Adam      & 200    & 0.98089907 & 6.32291809 & 665.1641919 \\
                12     & TriCNN   & MAE & 3 & 2 & (9, 7)          & {[}9, …, 5{]}  & RmsProp   & 250    & 0.98095925 & 6.72097815 & 669.8856237 \\
                13     & TriCNN   & MAE & 3 & 1 & (7, 6, 5, 4)    & {[}9, …, 5{]}  & RmsProp   & 200    & 0.98002879 & 7.62788323 & 684.7595471 \\
                $\sim$ & RF       &     &   &   &                 &                &           &        & 0.9803076  & 4.76701531 & 688.0001262 \\
                14     & TriCNN   & MAE & 2 & 1 & (9, 7)          & {[}11, …, 6{]} & Adam      & 150    & 0.9793459  & 6.519971   & 703.0615545 \\
                15     & TriCNN   & MAE & 3 & 2 & (9, 7)          & {[}9, …, 5{]}  & Adam      & 100    & \multicolumn{1}{l}{0.97782539} & \multicolumn{1}{l}{8.23651529} & 754.5381605 \\
                16     & TriCNN   & MAE & 3 & 2 & (9, 7, 6, 5, 4) & {[}9, …, 4{]}  & Adam      & 100    & 0.97748578 & 7.30871799 & 757.4994833 \\
                17     & TriCNN   & MAE & 3 & 2 & (9, 7, 6, 5, 4) & {[}9, …, 4{]}  & Adam      & 150    & 0.97726148 & 6.65855022 & 772.0747562 \\
                18     & TriCNN   & MAE & 3 & 1 & (7, 6, 5, 4)    & {[}9, …, 5{]}  & RmsProp   & 250    & 0.97665471 & 7.86703389 & \multicolumn{1}{l}{775.8960386} \\
                19     & TriCNN   & MAE & 3 & 2 & (9, 7)          & {[}9, …, 5{]}  & RmsProp   & 200    & 0.97613855 & 7.72461632 & 807.1294185 \\
                20     & TriCNN   & MAE & 3 & 2 & (9, 7)          & {[}9, …, 5{]}  & RmsProp   & 300    & 0.97636563 & 6.91501173 & 816.7881606 \\
                48     & TriMLP   & MAE &   &   &                 & {[}11, …, 6{]} & Adam      & 250    & \multicolumn{1}{l}{0.97347275} & \multicolumn{1}{l}{9.12443258} & \multicolumn{1}{l}{906.1439402} \\
                135    & Residual & MAE & \multicolumn{2}{l}{Number of Superblocks =   (2, 5, 5, 2)} & \multicolumn{1}{l}{((6, 6, 8), (7, 7, 9), (8, 8, 10), (9, 9, 11))} & 1 & RmsProp   & 250    & 0.95007233 & 10.595069 & 1006.134564 \\
                39     & LR       &     &   &   &                 &                &           &        & 0.52639158 & 82.4596122 & 15761.16107 \\
                40     & SVR      &     &   &   &                 &                &           &        & -0.0045634 & 113.749207 & 33448.30886 \\        
            \bottomrule
        \end{tabular}
    }
    \resizebox{1.5\columnwidth}{!}{* TriCNN = Trapezium-shaped CNN, RF = Random Forest Regression, TriMLP = Trapezium-shaped MPL, LR = Linear Regression, SVR = Support Vector Regression}
\end{table*}

We present the results of our model training. All results represent the average of the five different splits of the dataset. Tables \ref{tab:best20} and \ref{tab:best20mse} present the top performing models when sorted by $R^2$ and MSE. 

\subsection{Baseline Models}
\begin{figure*}[t]
  \centering
  \includegraphics[width=\linewidth]{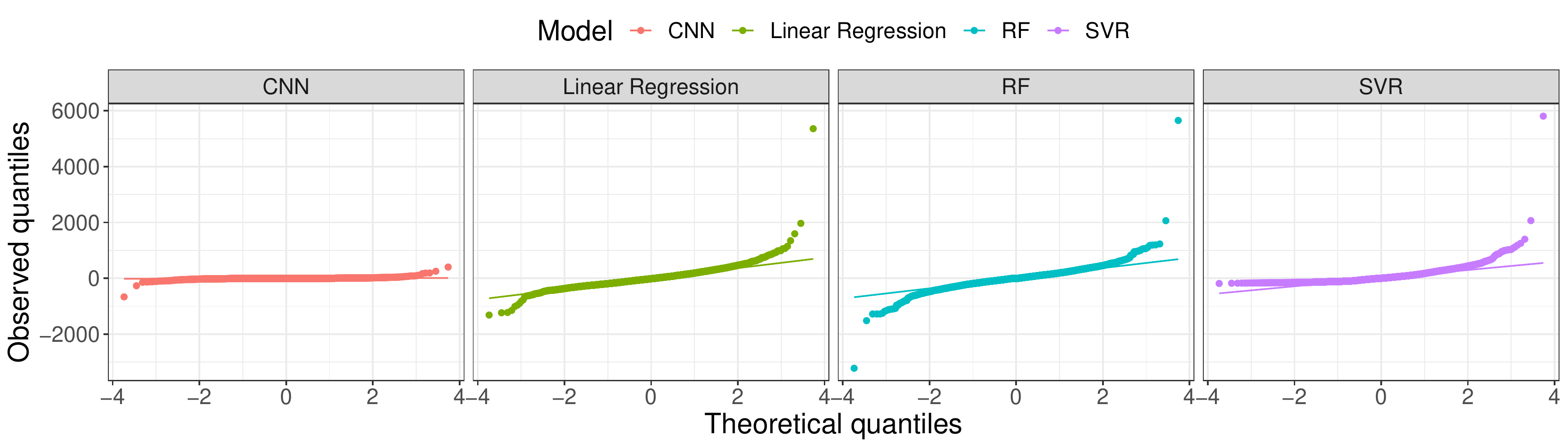}
  \caption{Q-Q Plots for residuals of the best performing CNN, Linear Regression, Random Forest and SVM models}
  \label{fig:qq}
\end{figure*}

\begin{figure}[t]
  \centering
  \includegraphics[width=\linewidth]{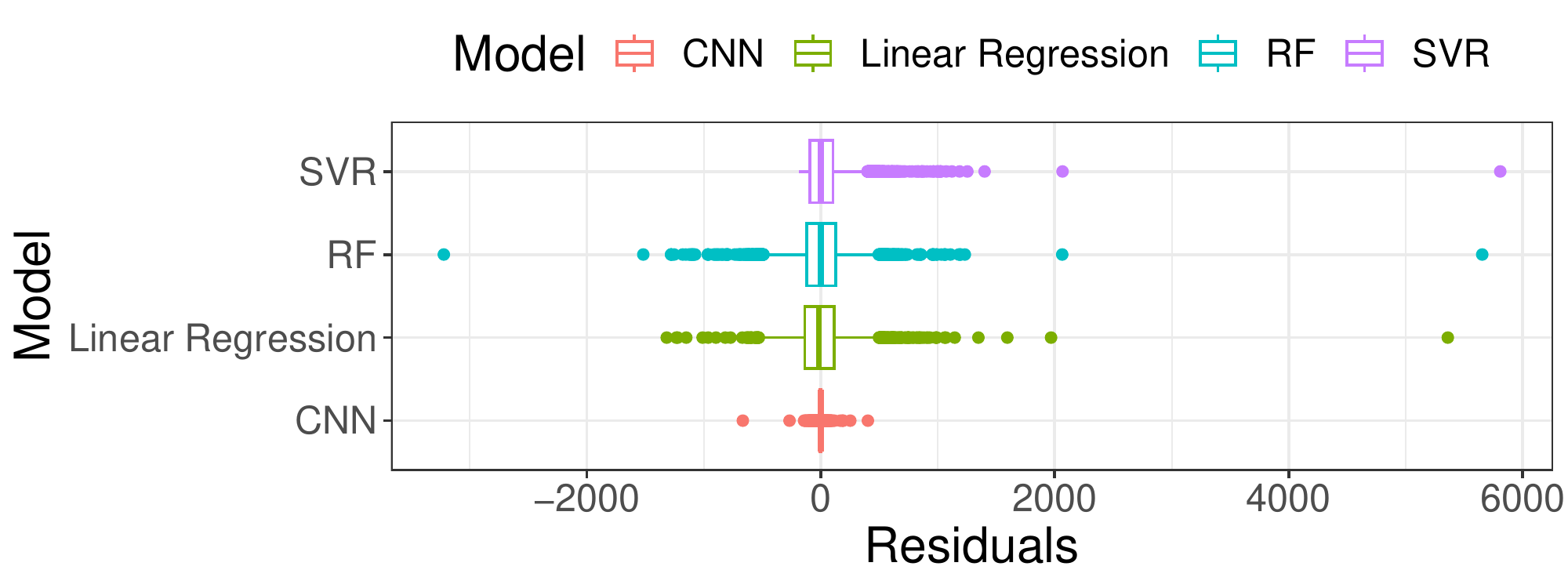}
  \caption{Magnitude of residuals, by method.}
  \label{fig:boxbymethod}
\end{figure}

We first evaluate our baseline cases. Both Linear Regression ($R^2 = 0.526$, $MSE=15761.2$) and Support Vector Regression  ($R^2 = -0.004$, $MSE=33448.31$) performed poorly. Figure~\ref{fig:qq} shows Quantile-Quantile plots for the residuals of the top-performing model of each type; CNN, Linear Regression, Random Forest and SVR. We observe that the CNN models exhibit preferable behaviour at both extremes. Meanwhile, linear regression and Random Forest models exhibit larger residuals for lower quantiles. Finally, Linear Regression, Random Forest and SVR exhibit large variances for high performance machines in the dataset. Figure~\ref{fig:boxbymethod} shows the magnitude of residuals for each methods. We observe our CNN based approach exhibits preferable behaviour to prior approaches.

\subsection{Deep Learning Models}
For MLP networks, the Trapezium networks offered highest performance, achieving 45th position for $R^2$ and 48th position for MSE. The performance of MLP networks were typically not competitive with CNN-based approaches, so we do not discuss them further. Meanwhile, CNN networks dominate the top 12 and 13 positions for $R^2$ and MSE respectively.

We hypothesised that residual-inspired approaches would perform favourably in our case, due to their strong performance in other domains, however, this is not borne in our findings. Residual-inspired approaches only acheived positions of 159th by $R^2$ and 135th by MSE. These models are more complex to engineer, and require more time to train, and provided little benefit in our case.

\begin{figure}[t]
  \centering
  \includegraphics[width=\linewidth]{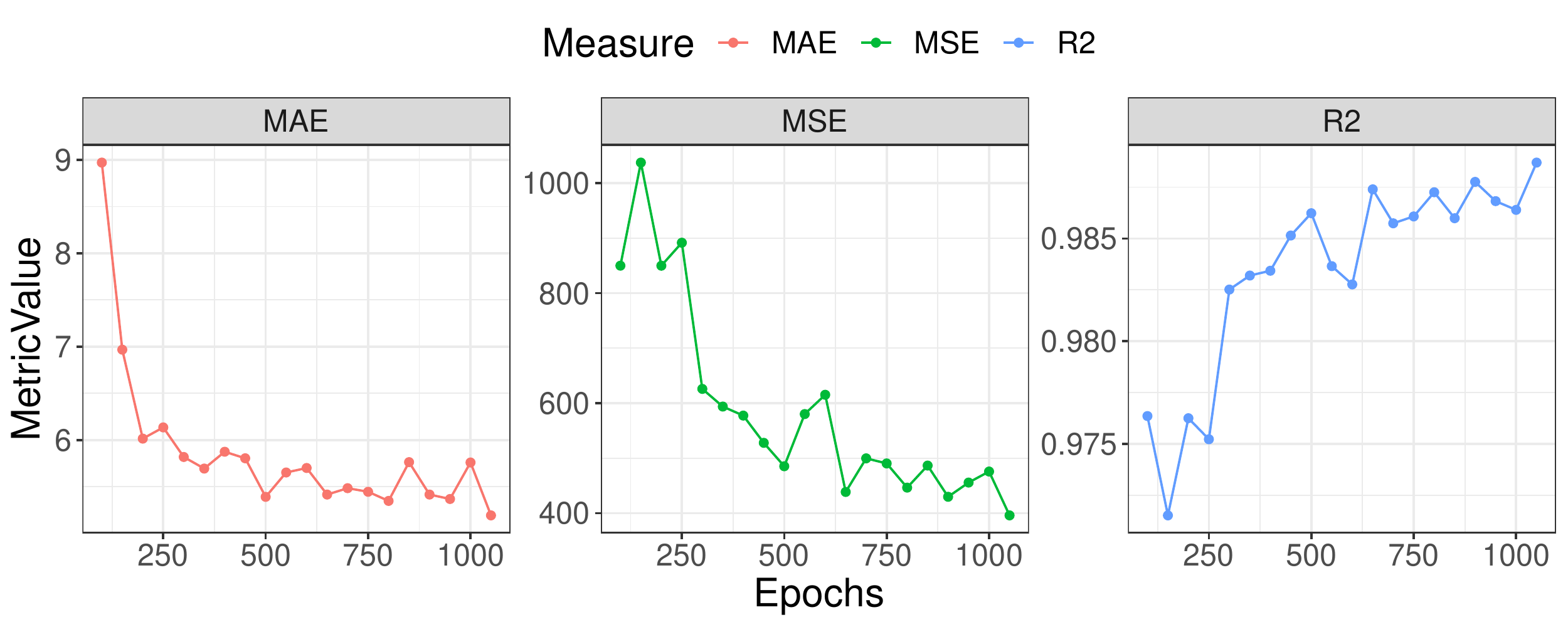}
  \caption{Performance for top CNN model, by training epoch.}
  \label{fig:epoch}
\end{figure}

We now summarise other design choices:
\begin{description}
    \item[Optimizer: ] Consistent with prior research, Adam generally performed best, though RMSprop is a strong contender.
    \item[Loss function: ] In all cases MAE produced the best results. Somewhat surprising when the overall metric is MSE.
    \item[Activation Function: ] Sigmoid produced results for the smaller architectures; however, the results were either NaN or negative. On the other hand, the results of the tanh activation function could not pass 0.01 in terms of R2. 
    \item[Stride Size: ] In most cases ($R^2$ and MSE) having a stride size of one and two are best. 
    \item[Kernel Size: ] The best kernel size is three for the top result though this is not consistent over all results. 
    \item[Training Epochs: ] Figure~\ref{fig:epoch} shows the impact of training epochs on performance for our top model, measured by MAE, MSE and $R^2$. The epoch count for stopping our training was determined empirically.
\end{description}

\subsection{Evaluation of top cases}
Here, we evaluate the performance of the top four networks by $R^2$ and MSE. As seen in Table~\ref{tab:best20} and~\ref{tab:best20mse}, four models which are trapezium-shaped CNNs that are optimised by Adam dominate both tables. All cases of the first-ranked model are in the top ten - 1st, 4th, 8th, and 10th in Table~\ref{tab:best20} and 1st, 4th, 8th, and 11th in Table~\ref{tab:best20mse}. However, when the table is examined closely, we see that the increase in the number of epochs does not result positively for this model and a fluctuation in $R^2$ performance is observed. At this point, it is important to compare other metrics. If we sort all epoch setups of the first-ranked model by MAE, we see similar fluctuations in performance. However, considering the MSE values, although there is no significant difference between each model, the first-ranked configuration seems to be more resistant to outliers. As a result, it is expected that an increase/decrease in the number of epochs affects the performance increase/decrease, but counter-intuitively, no pattern is obtained in this case. Moreover, the same argument is valid for the second-ranked model in both tables. Furthermore, an increase in the quantity of convolutional layers does not always translate into an improvement in performance.

\section{Threats to Validity}
\label{ttv}
Here, we introduce the limitations of this work, and highlight threats to validity arising from these. We structure our approach based on similar initiatives in the systems performance literature (e.g.,~\cite{eismann2022case}) and the approach of Wohlin \textit{et al.}~\cite{wohlin2012experimentation}.

\begin{description}
\item[L1 Single benchmark dataset] This study uses only data from SPEC CPU 2017 retrieved on 10 September 2022.
\item[L2 Single expert for data cleaning] Data cleaning processes were developed by a single expert researcher.
\end{description}

We now consider the implication of these limitations in terms of~\emph{construct},~\emph{internal} and~\emph{external} validity.

\textbf{Construct Validity} This work concerns the prediction of performance results. Further work could have also evaluate whether predictive performance holds for columns \emph{Energy Peak Result} and \emph{Energy Base Result} from the dataset. 

\textbf{Internal Validity} As highlighted in Section \ref{subse:data-clensing}, our work involved cleaning data for it to be amenable to analysis and machine learning. The development of the cleaning processes were undertaken by a single expert researcher (\textbf{Limitation L2}), leaving the opportunity for misinterpretation of the datasets. To mitigate this impact, the processes undertaken were well documented, and the process was audited by two further researchers. Code to automate data cleaning is made available to the community.

\textbf{External Validity} Our experiment considers data from a single benchmark, SPEC CPU 2017 (\textbf{Limitation L1}), which may limit the generalisability of our findings. While our experiments were conducted for just one benchmark, our methodology is applicable to performance benchmarks more broadly. Further research is required to understand the extent to which our methods are effective for other workloads; we make this possible by providing our data and models for reproduction by other researchers.

\textbf{Reproducibility} We have made all our code and data, including the results of the training of all the networks available\footnote{\url{https://github.com/cengizmehmet/BenchmarkNets}}.

\section{Conclusion}
\label{conc}
This work has considered the extent to which it is possibly to predict benchmark results for previously untested hardware configurations. We have specifically focused on the potential of using Deep Network approaches to capture the non-linear relationships present in the data. Our study has centred around the SPEC CPU 2017 dataset.

We investigated three deep network types, MLPs, CNNs, and an architecture of CNNs which is ResNet. After comprehensive studies, the models we offer excel at predicting the performance of a given system. While the $R^2$ values are between approx. 0.945 and 0.985, MAEs are between approx. 13 and 3.2. Secondly, it is discovered that convolutional layers can more efficiently predict our tabular data. This can be seen by examining the performance gain observed when adding convolutional layers to MLPs. Another finding of our paper involves demonstrating the effectiveness of residual blocks as opposed to simple convolutional layers. Our results indicate that while increasing the number of convolutional layers can offer promising results, the use of residual blocks leads to better performance overall.

This study is an indication and a starting point that deep neural networks can be trained on existing benchmark datasets to predict performance. However, we believe there are of many areas of future work. One avenue of future research would be to extend the application by taking advantage of more powerful neural network architectures with innovative feature aggregating modules or perhaps a higher parameter and layer count. Another direction of research would include exploring the effects of transfer learning, whereby the performance prediction system can be pre-trained on a larger proxy dataset to boost the performance after a subsequent and carefully designed fine-tuning process on the benchmark dataset. The use of synthetic data along with domain adaptation techniques can also lead to better performance and possibly steer the abilities of the model towards the desired outcome considering real-world data distributions. The method of procedurally generating the synthetic data in a meaningful manner that can benefit the training of neural network, perhaps in an end-to-end fashion, can also be an interesting area to investigate in a future work.

\bibliographystyle{ACM-Reference-Format}
\bibliography{sample-base.bbl}


\begin{thebibliography}{36}


\ifx \showCODEN    \undefined \def \showCODEN     #1{\unskip}     \fi
\ifx \showDOI      \undefined \def \showDOI       #1{#1}\fi
\ifx \showISBNx    \undefined \def \showISBNx     #1{\unskip}     \fi
\ifx \showISBNxiii \undefined \def \showISBNxiii  #1{\unskip}     \fi
\ifx \showISSN     \undefined \def \showISSN      #1{\unskip}     \fi
\ifx \showLCCN     \undefined \def \showLCCN      #1{\unskip}     \fi
\ifx \shownote     \undefined \def \shownote      #1{#1}          \fi
\ifx \showarticletitle \undefined \def \showarticletitle #1{#1}   \fi
\ifx \showURL      \undefined \def \showURL       {\relax}        \fi
\providecommand\bibfield[2]{#2}
\providecommand\bibinfo[2]{#2}
\providecommand\natexlab[1]{#1}
\providecommand\showeprint[2][]{arXiv:#2}

\bibitem[3DMark(2022)]%
        {3dmark}
\bibfield{author}{\bibinfo{person}{3DMark}.} \bibinfo{year}{2022}\natexlab{}.
\newblock \bibinfo{booktitle}{\emph{3DMark Benchmarks}}.
\newblock
\urldef\tempurl%
\url{https://www.3dmark.com/}
\showURL{%
\tempurl}
\newblock
\shownote{Last accessed 7 October 2022}.


\bibitem[Anand and Kodali(2008)]%
        {anand2008benchmarking}
\bibfield{author}{\bibinfo{person}{Gurumurthy Anand} {and}
  \bibinfo{person}{Rambabu Kodali}.} \bibinfo{year}{2008}\natexlab{}.
\newblock \showarticletitle{Benchmarking the benchmarking models}.
\newblock \bibinfo{journal}{\emph{Benchmarking: An international journal}}
  \bibinfo{volume}{15}, \bibinfo{number}{3} (\bibinfo{year}{2008}),
  \bibinfo{pages}{257--291}.
\newblock
\urldef\tempurl%
\url{https://doi.org/10.1108/14635770810876593}
\showDOI{\tempurl}


\bibitem[Ardalani et~al\mbox{.}(2015)]%
        {ardalani}
\bibfield{author}{\bibinfo{person}{Newsha Ardalani}, \bibinfo{person}{Clint
  Lestourgeon}, \bibinfo{person}{Karthikeyan Sankaralingam}, {and}
  \bibinfo{person}{Xiaojin Zhu}.} \bibinfo{year}{2015}\natexlab{}.
\newblock \showarticletitle{Cross-Architecture Performance Prediction (XAPP)
  Using CPU Code to Predict GPU Performance}. In
  \bibinfo{booktitle}{\emph{Proceedings of the 48th International Symposium on
  Microarchitecture}} (Waikiki, Hawaii) \emph{(\bibinfo{series}{MICRO-48})}.
  \bibinfo{publisher}{Association for Computing Machinery},
  \bibinfo{address}{New York, NY, USA}, \bibinfo{pages}{725–737}.
\newblock
\showISBNx{9781450340342}
\urldef\tempurl%
\url{https://doi.org/10.1145/2830772.2830780}
\showDOI{\tempurl}


\bibitem[baosenguo(2020)]%
        {kaggle_1d}
\bibfield{author}{\bibinfo{person}{baosenguo}.}
  \bibinfo{year}{2020}\natexlab{}.
\newblock \bibinfo{title}{Mechanisms of Action (MoA) Prediction: 2nd Place
  Solution - with 1D-CNN}.
\newblock
\newblock


\bibitem[Buturovi{\'c} and Miljkovi{\'c}(2020)]%
        {Butur}
\bibfield{author}{\bibinfo{person}{Ljubomir Buturovi{\'c}} {and}
  \bibinfo{person}{Dejan Miljkovi{\'c}}.} \bibinfo{year}{2020}\natexlab{}.
\newblock \showarticletitle{A novel method for classification of tabular data
  using convolutional neural networks}.
\newblock \bibinfo{journal}{\emph{bioRxiv}} (\bibinfo{year}{2020}).
\newblock
\urldef\tempurl%
\url{https://doi.org/10.1101/2020.05.02.074203}
\showDOI{\tempurl}
\showeprint{https://www.biorxiv.org/content/early/2020/05/03/2020.05.02.074203.full.pdf}


\bibitem[Camp(2006)]%
        {camp1989benchmarking}
\bibfield{author}{\bibinfo{person}{Robert~C Camp}.}
  \bibinfo{year}{2006}\natexlab{}.
\newblock \bibinfo{booktitle}{\emph{Benchmarking: the search for industry best
  practices that lead to superior performance} (\bibinfo{edition}{1st.} ed.)}.
\newblock \bibinfo{publisher}{Productivity Press, NY}.
\newblock


\bibitem[Ein-Dor and Feldmesser(1987)]%
        {ein-dor}
\bibfield{author}{\bibinfo{person}{Phillip Ein-Dor} {and}
  \bibinfo{person}{Jacob Feldmesser}.} \bibinfo{year}{1987}\natexlab{}.
\newblock \showarticletitle{Attributes of the Performance of Central Processing
  Units: A Relative Performance Prediction Model}.
\newblock \bibinfo{journal}{\emph{Commun. ACM}} \bibinfo{volume}{30},
  \bibinfo{number}{4} (\bibinfo{date}{apr} \bibinfo{year}{1987}),
  \bibinfo{pages}{308–317}.
\newblock
\showISSN{0001-0782}
\urldef\tempurl%
\url{https://doi.org/10.1145/32232.32234}
\showDOI{\tempurl}


\bibitem[Eismann et~al\mbox{.}(2022)]%
        {eismann2022case}
\bibfield{author}{\bibinfo{person}{Simon Eismann}, \bibinfo{person}{Diego~Elias
  Costa}, \bibinfo{person}{Lizhi Liao}, \bibinfo{person}{Cor-Paul Bezemer},
  \bibinfo{person}{Weiyi Shang}, \bibinfo{person}{Andr{\'e} van Hoorn}, {and}
  \bibinfo{person}{Samuel Kounev}.} \bibinfo{year}{2022}\natexlab{}.
\newblock \showarticletitle{A case study on the stability of performance tests
  for serverless applications}.
\newblock \bibinfo{journal}{\emph{Journal of Systems and Software}}
  \bibinfo{volume}{189} (\bibinfo{year}{2022}), \bibinfo{pages}{111294}.
\newblock


\bibitem[Eyerman et~al\mbox{.}(2009)]%
        {eyerman_mech}
\bibfield{author}{\bibinfo{person}{Stijn Eyerman}, \bibinfo{person}{Lieven
  Eeckhout}, \bibinfo{person}{Tejas Karkhanis}, {and} \bibinfo{person}{James~E.
  Smith}.} \bibinfo{year}{2009}\natexlab{}.
\newblock \showarticletitle{A Mechanistic Performance Model for Superscalar
  Out-of-Order Processors}.
\newblock \bibinfo{journal}{\emph{ACM Trans. Comput. Syst.}}
  \bibinfo{volume}{27}, \bibinfo{number}{2}, Article \bibinfo{articleno}{3}
  (\bibinfo{date}{may} \bibinfo{year}{2009}), \bibinfo{numpages}{37}~pages.
\newblock
\showISSN{0734-2071}
\urldef\tempurl%
\url{https://doi.org/10.1145/1534909.1534910}
\showDOI{\tempurl}


\bibitem[Geekbench(2004)]%
        {geek}
\bibfield{author}{\bibinfo{person}{Geekbench}.}
  \bibinfo{year}{2004}\natexlab{}.
\newblock \bibinfo{booktitle}{\emph{Geekbench 5}}.
\newblock
\urldef\tempurl%
\url{https://browser.geekbench.com/}
\showURL{%
\tempurl}
\newblock
\shownote{Last accessed 7 February 2022}.


\bibitem[Glorot and Bengio(2010)]%
        {glorot}
\bibfield{author}{\bibinfo{person}{Xavier Glorot} {and} \bibinfo{person}{Yoshua
  Bengio}.} \bibinfo{year}{2010}\natexlab{}.
\newblock \showarticletitle{Understanding the difficulty of training deep
  feedforward neural networks}. In \bibinfo{booktitle}{\emph{Proceedings of the
  Thirteenth International Conference on Artificial Intelligence and
  Statistics}} \emph{(\bibinfo{series}{Proceedings of Machine Learning
  Research}, Vol.~\bibinfo{volume}{9})},
  \bibfield{editor}{\bibinfo{person}{Yee~Whye Teh} {and} \bibinfo{person}{Mike
  Titterington}} (Eds.). \bibinfo{publisher}{PMLR}, \bibinfo{address}{Chia
  Laguna Resort, Sardinia, Italy}, \bibinfo{pages}{249--256}.
\newblock
\urldef\tempurl%
\url{https://proceedings.mlr.press/v9/glorot10a.html}
\showURL{%
\tempurl}


\bibitem[He et~al\mbox{.}(2015)]%
        {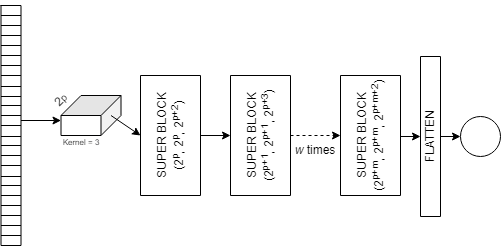}
\bibfield{author}{\bibinfo{person}{Kaiming He}, \bibinfo{person}{Xiangyu
  Zhang}, \bibinfo{person}{Shaoqing Ren}, {and} \bibinfo{person}{Jian Sun}.}
  \bibinfo{year}{2015}\natexlab{}.
\newblock \showarticletitle{Deep Residual Learning for Image Recognition}.
\newblock \bibinfo{journal}{\emph{CoRR}}  \bibinfo{volume}{abs/1512.03385}
  (\bibinfo{year}{2015}).
\newblock
\showeprint[arXiv]{1512.03385}
\urldef\tempurl%
\url{http://arxiv.org/abs/1512.03385}
\showURL{%
\tempurl}


\bibitem[Hugue(2022)]%
        {umd}
\bibfield{author}{\bibinfo{person}{Michelle~M. Hugue}.}
  \bibinfo{year}{2022}\natexlab{}.
\newblock \bibinfo{title}{Lecture notes in Computer Systems Architecture}.
\newblock
\newblock


\bibitem[\"{I}pek et~al\mbox{.}(2006)]%
        {engin}
\bibfield{author}{\bibinfo{person}{Engin \"{I}pek}, \bibinfo{person}{Sally~A.
  McKee}, \bibinfo{person}{Rich Caruana}, \bibinfo{person}{Bronis~R. de
  Supinski}, {and} \bibinfo{person}{Martin Schulz}.}
  \bibinfo{year}{2006}\natexlab{}.
\newblock \showarticletitle{Efficiently Exploring Architectural Design Spaces
  via Predictive Modeling}. In \bibinfo{booktitle}{\emph{Proceedings of the
  12th International Conference on Architectural Support for Programming
  Languages and Operating Systems}} (San Jose, California, USA)
  \emph{(\bibinfo{series}{ASPLOS XII})}. \bibinfo{publisher}{Association for
  Computing Machinery}, \bibinfo{address}{New York, NY, USA},
  \bibinfo{pages}{195–206}.
\newblock
\showISBNx{1595934510}
\urldef\tempurl%
\url{https://doi.org/10.1145/1168857.1168882}
\showDOI{\tempurl}


\bibitem[Jiang et~al\mbox{.}(2013)]%
        {Jiang}
\bibfield{author}{\bibinfo{person}{Chuntao Jiang}, \bibinfo{person}{Zhibin Yu},
  \bibinfo{person}{Hai Jin}, \bibinfo{person}{Chengzhong Xu},
  \bibinfo{person}{Lieven Eeckhout}, \bibinfo{person}{Wim Heirman},
  \bibinfo{person}{Trevor~E. Carlson}, {and} \bibinfo{person}{Xiaofei Liao}.}
  \bibinfo{year}{2013}\natexlab{}.
\newblock \showarticletitle{PCantorSim: Accelerating Parallel Architecture
  Simulation through Fractal-Based Sampling}.
\newblock \bibinfo{journal}{\emph{ACM Trans. Archit. Code Optim.}}
  \bibinfo{volume}{10}, \bibinfo{number}{4}, Article \bibinfo{articleno}{49}
  (\bibinfo{date}{dec} \bibinfo{year}{2013}), \bibinfo{numpages}{24}~pages.
\newblock
\showISSN{1544-3566}
\urldef\tempurl%
\url{https://doi.org/10.1145/2541228.2555305}
\showDOI{\tempurl}


\bibitem[Justus et~al\mbox{.}(2018)]%
        {justus}
\bibfield{author}{\bibinfo{person}{Daniel Justus}, \bibinfo{person}{John
  Brennan}, \bibinfo{person}{Stephen Bonner}, {and}
  \bibinfo{person}{Andrew~Stephen McGough}.} \bibinfo{year}{2018}\natexlab{}.
\newblock \showarticletitle{Predicting the Computational Cost of Deep Learning
  Models}. In \bibinfo{booktitle}{\emph{2018 IEEE International Conference on
  Big Data (Big Data)}}. \bibinfo{pages}{3873--3882}.
\newblock
\urldef\tempurl%
\url{https://doi.org/10.1109/BigData.2018.8622396}
\showDOI{\tempurl}


\bibitem[Keras(2022)]%
        {kerasInit}
\bibfield{author}{\bibinfo{person}{Keras}.} \bibinfo{year}{2022}\natexlab{}.
\newblock \bibinfo{booktitle}{\emph{Layer weight initializers}}.
\newblock
\urldef\tempurl%
\url{https://keras.io/api/layers/initializers/}
\showURL{%
\tempurl}
\newblock
\shownote{Last accessed 15 August 2022}.


\bibitem[Kingma and Ba(2014)]%
        {adam}
\bibfield{author}{\bibinfo{person}{Diederik~P. Kingma} {and}
  \bibinfo{person}{Jimmy Ba}.} \bibinfo{year}{2014}\natexlab{}.
\newblock \bibinfo{title}{Adam: A Method for Stochastic Optimization}.
\newblock
\newblock
\urldef\tempurl%
\url{https://doi.org/10.48550/ARXIV.1412.6980}
\showDOI{\tempurl}


\bibitem[Lee(2006)]%
        {benjamin_spec}
\bibfield{author}{\bibinfo{person}{Benjamin Lee}.}
  \bibinfo{year}{2006}\natexlab{}.
\newblock \showarticletitle{An architectural assessment of SPEC CPU benchmark
  relevance}.
\newblock \bibinfo{journal}{\emph{Harvard University, Cambridge, MA, Tech. Rep.
  TR-02-06}} (\bibinfo{year}{2006}).
\newblock


\bibitem[Lee and Brooks(2006)]%
        {benjamin}
\bibfield{author}{\bibinfo{person}{Benjamin~C. Lee} {and}
  \bibinfo{person}{David~M. Brooks}.} \bibinfo{year}{2006}\natexlab{}.
\newblock \showarticletitle{Accurate and Efficient Regression Modeling for
  Microarchitectural Performance and Power Prediction}
  \emph{(\bibinfo{series}{ASPLOS XII})}. \bibinfo{publisher}{Association for
  Computing Machinery}, \bibinfo{address}{New York, NY, USA},
  \bibinfo{numpages}{10}~pages.
\newblock
\showISBNx{1595934510}
\urldef\tempurl%
\url{https://doi.org/10.1145/1168857.1168881}
\showDOI{\tempurl}


\bibitem[Li et~al\mbox{.}(2011)]%
        {li_cloud}
\bibfield{author}{\bibinfo{person}{Ang Li}, \bibinfo{person}{Xuanran Zong},
  \bibinfo{person}{Srikanth Kandula}, \bibinfo{person}{Xiaowei Yang}, {and}
  \bibinfo{person}{Ming Zhang}.} \bibinfo{year}{2011}\natexlab{}.
\newblock \showarticletitle{CloudProphet: Towards Application Performance
  Prediction in Cloud} \emph{(\bibinfo{series}{SIGCOMM '11})}.
  \bibinfo{publisher}{Association for Computing Machinery},
  \bibinfo{address}{New York, NY, USA}, \bibinfo{pages}{426–427}.
\newblock
\showISBNx{9781450307970}
\urldef\tempurl%
\url{https://doi.org/10.1145/2018436.2018502}
\showDOI{\tempurl}


\bibitem[Lilja(2000)]%
        {cam_bench}
\bibfield{author}{\bibinfo{person}{David~J. Lilja}.}
  \bibinfo{year}{2000}\natexlab{}.
\newblock \bibinfo{title}{Measuring computer performance: A practitioner's
  guide}.
\newblock
\newblock


\bibitem[Lopez et~al\mbox{.}(2018)]%
        {lopez2018}
\bibfield{author}{\bibinfo{person}{Leonardo Lopez}, \bibinfo{person}{Michael
  Guynn}, {and} \bibinfo{person}{Meiliu Lu}.} \bibinfo{year}{2018}\natexlab{}.
\newblock \showarticletitle{Predicting Computer Performance Based on Hardware
  Configuration Using Multiple Neural Networks}. In
  \bibinfo{booktitle}{\emph{ICMLA}}. \bibinfo{pages}{824--827}.
\newblock
\urldef\tempurl%
\url{https://doi.org/10.1109/ICMLA.2018.00132}
\showDOI{\tempurl}


\bibitem[Maire et~al\mbox{.}(2005)]%
        {maire2005typology}
\bibfield{author}{\bibinfo{person}{Jean-Luc Maire}, \bibinfo{person}{Vincent},
  \bibinfo{person}{Pillet Bronet}, {and} \bibinfo{person}{Maurice Pillet}.}
  \bibinfo{year}{2005}\natexlab{}.
\newblock \showarticletitle{A typology of “best practices” for a
  benchmarking process}.
\newblock \bibinfo{journal}{\emph{Benchmarking: An international journal}}
  \bibinfo{volume}{12}, \bibinfo{number}{1} (\bibinfo{year}{2005}),
  \bibinfo{pages}{45--60}.
\newblock
\urldef\tempurl%
\url{https://doi.org/10.1108/14635770510582907}
\showDOI{\tempurl}


\bibitem[Ozisikyilmaz et~al\mbox{.}(2008)]%
        {berkin2008}
\bibfield{author}{\bibinfo{person}{Berkin Ozisikyilmaz},
  \bibinfo{person}{Gokhan Memik}, {and} \bibinfo{person}{Alok Choudhary}.}
  \bibinfo{year}{2008}\natexlab{}.
\newblock \showarticletitle{Machine Learning Models to Predict Performance of
  Computer System Design Alternatives}. In \bibinfo{booktitle}{\emph{2008 37th
  International Conference on Parallel Processing}}. \bibinfo{pages}{495--502}.
\newblock
\urldef\tempurl%
\url{https://doi.org/10.1109/ICPP.2008.36}
\showDOI{\tempurl}


\bibitem[PassMark(2021)]%
        {passmark}
\bibfield{author}{\bibinfo{person}{PassMark}.} \bibinfo{year}{2021}\natexlab{}.
\newblock \bibinfo{booktitle}{\emph{CPU Benchmarks}}.
\newblock
\urldef\tempurl%
\url{https://www.cpubenchmark.net/}
\showURL{%
\tempurl}
\newblock
\shownote{Last accessed 7 February 2022}.


\bibitem[Sarmento(2022)]%
        {kendall}
\bibfield{author}{\bibinfo{person}{David Sarmento}.}
  \bibinfo{year}{2022}\natexlab{}.
\newblock \bibinfo{booktitle}{\emph{Chapter 22: Correlation Types and When to
  Use Them}}.
\newblock


\bibitem[Singh et~al\mbox{.}(2007)]%
        {karan}
\bibfield{author}{\bibinfo{person}{Karan Singh}, \bibinfo{person}{Engin İpek},
  \bibinfo{person}{Sally~A. McKee}, \bibinfo{person}{Bronis~R. de Supinski},
  \bibinfo{person}{Martin Schulz}, {and} \bibinfo{person}{Rich Caruana}.}
  \bibinfo{year}{2007}\natexlab{}.
\newblock \showarticletitle{Predicting parallel application performance via
  machine learning approaches}.
\newblock \bibinfo{journal}{\emph{Concurrency and Computation: Practice and
  Experience}} \bibinfo{volume}{19}, \bibinfo{number}{17}
  (\bibinfo{year}{2007}), \bibinfo{pages}{2219--2235}.
\newblock
\urldef\tempurl%
\url{https://doi.org/10.1002/cpe.1171}
\showDOI{\tempurl}
\showeprint{https://onlinelibrary.wiley.com/doi/pdf/10.1002/cpe.1171}


\bibitem[Spec(1988)]%
        {spec}
\bibfield{author}{\bibinfo{person}{Spec}.} \bibinfo{year}{1988}\natexlab{}.
\newblock \bibinfo{booktitle}{\emph{SPEC}}.
\newblock
\urldef\tempurl%
\url{https://www.spec.org/}
\showURL{%
\tempurl}
\newblock
\shownote{Last accessed 7 February 2022}.


\bibitem[Tousi and Luján(2022)]%
        {tousi}
\bibfield{author}{\bibinfo{person}{Ashkan Tousi} {and} \bibinfo{person}{Mikel
  Luján}.} \bibinfo{year}{2022}\natexlab{}.
\newblock \showarticletitle{Comparative Analysis of Machine Learning Models for
  Performance Prediction of the SPEC Benchmarks}.
\newblock \bibinfo{journal}{\emph{IEEE Access}}  \bibinfo{volume}{10}
  (\bibinfo{year}{2022}), \bibinfo{pages}{11994--12011}.
\newblock
\urldef\tempurl%
\url{https://doi.org/10.1109/ACCESS.2022.3142240}
\showDOI{\tempurl}


\bibitem[Upadhyay et~al\mbox{.}(2022)]%
        {Upadhyay}
\bibfield{author}{\bibinfo{person}{Navin~Mani Upadhyay},
  \bibinfo{person}{Ravi~Shankar Singh}, {and} \bibinfo{person}{Shri~Prakash
  Dwivedi}.} \bibinfo{year}{2022}\natexlab{}.
\newblock \showarticletitle{Prediction of multicore CPU performance through
  parallel data mining on public datasets}.
\newblock \bibinfo{journal}{\emph{Displays}}  \bibinfo{volume}{71}
  (\bibinfo{year}{2022}), \bibinfo{pages}{102112}.
\newblock
\showISSN{0141-9382}
\urldef\tempurl%
\url{https://doi.org/10.1016/j.displa.2021.102112}
\showDOI{\tempurl}


\bibitem[Van~den Steen et~al\mbox{.}(2015)]%
        {emp_per}
\bibfield{author}{\bibinfo{person}{Sam Van~den Steen}, \bibinfo{person}{Sander
  De~Pestel}, \bibinfo{person}{Moncef Mechri}, \bibinfo{person}{Stijn Eyerman},
  \bibinfo{person}{Trevor Carlson}, \bibinfo{person}{David Black-Schaffer},
  \bibinfo{person}{Erik Hagersten}, {and} \bibinfo{person}{Lieven Eeckhout}.}
  \bibinfo{year}{2015}\natexlab{}.
\newblock \showarticletitle{Micro-architecture independent analytical processor
  performance and power modeling}. In \bibinfo{booktitle}{\emph{2015 IEEE
  International Symposium on Performance Analysis of Systems and Software
  (ISPASS)}}. \bibinfo{pages}{32--41}.
\newblock
\urldef\tempurl%
\url{https://doi.org/10.1109/ISPASS.2015.7095782}
\showDOI{\tempurl}


\bibitem[Wohlin et~al\mbox{.}(2012)]%
        {wohlin2012experimentation}
\bibfield{author}{\bibinfo{person}{Claes Wohlin}, \bibinfo{person}{Per
  Runeson}, \bibinfo{person}{Martin H{\"o}st}, \bibinfo{person}{Magnus~C
  Ohlsson}, \bibinfo{person}{Bj{\"o}rn Regnell}, {and} \bibinfo{person}{Anders
  Wessl{\'e}n}.} \bibinfo{year}{2012}\natexlab{}.
\newblock \bibinfo{booktitle}{\emph{Experimentation in software engineering}}.
\newblock \bibinfo{publisher}{Springer Science \& Business Media}.
\newblock


\bibitem[Zeng and Martinez(2000)]%
        {cv}
\bibfield{author}{\bibinfo{person}{Xinchuan Zeng} {and}
  \bibinfo{person}{Tony~R. Martinez}.} \bibinfo{year}{2000}\natexlab{}.
\newblock \showarticletitle{Distribution-balanced stratified cross-validation
  for accuracy estimation}.
\newblock \bibinfo{journal}{\emph{Journal of Experimental \& Theoretical
  Artificial Intelligence}} \bibinfo{volume}{12}, \bibinfo{number}{1}
  (\bibinfo{year}{2000}), \bibinfo{pages}{1--12}.
\newblock
\urldef\tempurl%
\url{https://doi.org/10.1080/095281300146272}
\showDOI{\tempurl}


\bibitem[Zheng et~al\mbox{.}(2016)]%
        {Zheng_spec}
\bibfield{author}{\bibinfo{person}{Xinnian Zheng}, \bibinfo{person}{Lizy~K.
  John}, {and} \bibinfo{person}{Andreas Gerstlauer}.}
  \bibinfo{year}{2016}\natexlab{}.
\newblock \showarticletitle{Accurate phase-level cross-platform power and
  performance estimation}. In \bibinfo{booktitle}{\emph{2016 53nd ACM/EDAC/IEEE
  Design Automation Conference (DAC)}}. \bibinfo{pages}{1--6}.
\newblock
\urldef\tempurl%
\url{https://doi.org/10.1145/2897937.2897977}
\showDOI{\tempurl}


\bibitem[Zhu et~al\mbox{.}(2021)]%
        {tab_conv}
\bibfield{author}{\bibinfo{person}{Yitan Zhu}, \bibinfo{person}{Thomas
  Brettin}, \bibinfo{person}{Fangfang Xia}, \bibinfo{person}{Alexander Partin},
  \bibinfo{person}{Maulik Shukla}, \bibinfo{person}{Hyunseung Yoo},
  \bibinfo{person}{Yvonne~A. Evrard}, \bibinfo{person}{James~H. Doroshow},
  {and} \bibinfo{person}{Rick~L. Stevens}.} \bibinfo{year}{2021}\natexlab{}.
\newblock \showarticletitle{Converting tabular data into images for deep
  learning with convolutional neural networks}.
\newblock \bibinfo{journal}{\emph{Scientific Reports}} \bibinfo{volume}{11},
  \bibinfo{number}{1}, Article \bibinfo{articleno}{11325}
  (\bibinfo{year}{2021}), \bibinfo{numpages}{11}~pages.
\newblock
\urldef\tempurl%
\url{https://doi.org/10.1038/s41598-021-90923-y}
\showDOI{\tempurl}


\end{thebibliography}

\end{document}